\documentclass[runningheads]{llncs}
\usepackage[T1]{fontenc}
\usepackage{graphicx}
\usepackage{amsmath}
\usepackage{amssymb}
\usepackage{booktabs}
\usepackage{multirow}
\usepackage{xcolor}
\usepackage{tikz}
\usetikzlibrary{positioning,arrows.meta}
\usepackage{pgfplots}
\pgfplotsset{compat=1.17}

\begin{document}
\title{HAN-Mamba: Hierarchical Selective State Space Networks for Multi-Scale Financial Volatility Forecasting\thanks{This article is a revised and extended version of our paper presented at ICAART 2026~\cite{deaconu2026hierarchical}.}}
\titlerunning{Hierarchical Selective SSMs for Volatility Forecasting}
\author{Mihai Bogdan Deaconu\inst{1}\orcidID{0009-0003-9744-6091} \and
Ioan Daniel Pop\inst{1}\orcidID{0000-0002-3740-6579}}
\authorrunning{M.B. Deaconu and I.D. Pop}
\institute{Faculty of Mathematics and Computer Science, Babe\c{s}-Bolyai University,\\ Cluj-Napoca, Romania\\
\email{deaconu.mihai.bogdan@gmail.com, ioan.daniel.pop@ubbcluj.ro}}
\maketitle
\begin{abstract}
Short-horizon realized volatility forecasting requires the integration of market information that evolves at incompatible temporal resolutions, from second-level order book dynamics to weekly regime drift. Our conference work introduced HAN-T, a hierarchical architecture in which scale-specific Transformer encoders process short, mid, and long-horizon streams and a learned attention fuser weighs their contributions. This article replaces the quadratic attention encoders with selective state space (Mamba) encoders while retaining attention only in the fuser, where the input is a three-token set rather than a long sequence. The resulting hybrid, HAN-Mamba, summarizes each stream through a recurrent state whose input-dependent gating matches two structural properties of volatility: persistent but decaying memory and abrupt regime shifts. On the Optiver Realized Volatility Prediction benchmark under time-aware five-fold cross-validation, HAN-Mamba improves mean RMSPE over HAN-T (0.1942 vs. 0.1965) with 33\% fewer parameters. Its linear-time encoders further allow the high-frequency context to be extended from 60 to 240 buckets, reducing error to 0.1927 where the attention variant saturates, and support constant-time streaming updates at inference. Ablations attribute the gains to the encoder swap, confirm that the hierarchical prior transfers across sequence-model families, and show that the permutation-invariant attention fuser remains the correct mechanism for cross-scale integration.

\end{abstract}
\section{Introduction}
Realized volatility is the operative measure of short-term risk in modern electronic markets. Accurate forecasts over horizons of minutes drive option market making, execution scheduling, and intraday risk limits, and the data supporting such forecasts arrive as high-frequency order book and trade streams whose statistical structure spans several temporal scales at once. Volatility clusters at the daily scale, decays slowly at the weekly scale, and reacts within seconds to liquidity shocks, a combination that has long resisted single-resolution models~\cite{cont2001empirical,andersen2003modeling}.

In our conference paper~\cite{deaconu2026hierarchical} we addressed this multi-scale structure architecturally. The proposed HAN-T processes three time-synchronized streams, a high-frequency feature sequence together with daily and weekly volatility histories, through asymmetric scale-specific Transformer encoders, and integrates the resulting summary tokens with a lightweight attention fuser. Under time-aware cross-validation on the Optiver Realized Volatility Prediction (ORVP) benchmark, HAN-T outperformed econometric, gradient boosting, and flat Transformer baselines, and ablations isolated the contribution of both the hierarchy and the learned fusion.

\subsubsection{Limitations of Attention Encoders in This Setting.}
The encoders carry almost all of the sequence-processing burden in HAN-T, and self-attention is an expensive mechanism for that role. Its cost and activation memory grow quadratically with sequence length, which in practice caps the high-frequency context at 60 buckets and makes longer histories disproportionately costly. Order information must be injected through learned absolute positional embeddings, which are tied to a fixed training length and transfer poorly when the context is extended. At deployment time, each new 10-minute bucket forces a full recomputation over the entire window, since the encoder maintains no state between predictions. None of these costs purchases a needed capability: the encoder output is a single summary vector, not a token-level representation, so the all-pairs interaction structure that justifies attention is largely discarded at readout.

\subsubsection{Selective State Spaces as Sequence Encoders.}
Selective state space models~\cite{gu2023mamba} offer a sequence operator whose inductive bias aligns with volatility dynamics rather than merely approximating attention at lower cost. A discretized state space recurrence propagates a hidden state with multiplicative decay, the natural analogue of volatility persistence with fading memory~\cite{corsi2009simple}. The selection mechanism of Mamba makes the state transition input-dependent, so the model can retain information through calm periods and rapidly discount stale context when a liquidity shock arrives, an architectural counterpart of regime-dependent memory. Computation is linear in sequence length, inference reduces to a constant-time state update per new observation, and the hardware-aware scan keeps the operator fast on modern accelerators~\cite{gu2023mamba,dao2024transformers}.

\subsubsection{Our Approach.}
We propose HAN-Mamba, a hybrid revision of HAN-T that allocates each mechanism where its inductive bias is warranted. The three scale-specific encoders become stacks of Mamba blocks read out through their final recurrent state, which removes positional embeddings and the classification token entirely. The fuser remains a small Transformer: its input is a set of three scale summaries with no canonical ordering, so a causal scan would impose an arbitrary bias, while at length three the quadratic cost objection is vacuous and the attention weights stay directly interpretable. This selective placement follows the design logic of recent hybrid attention-SSM stacks~\cite{lieber2024jamba}, applied here at the granularity of architectural roles rather than interleaved layers. Our contributions are summarized as follows:

\begin{enumerate}
\item We introduce HAN-Mamba, a hierarchical hybrid in which selective state space encoders with final-state readout replace the Transformer encoders of the conference architecture, and we motivate the swap through the correspondence between selective forgetting and regime-dependent volatility memory (Sect.~\ref{sec:hanm}).
\item We give a set-versus-sequence analysis of cross-scale fusion, arguing that attention is the correct operator for the three-token fusion stage, and we validate the analysis with an ablation over four fusion designs, including order-sensitivity of recurrent fusers (Sect.~\ref{sec:ablations}).
\item We exploit the linear-time encoders to extend the high-frequency context from 60 to 240 buckets, a regime the conference architecture could not usefully reach, and we characterize accuracy and cost along this axis (Sects.~\ref{sec:context} and~\ref{sec:efficiency}).
\item We strengthen the econometric grounding of the framework by connecting the three-stream hierarchy to the heterogeneous-horizon cascade of the HAR model and to the stylized facts of volatility, positioning the architecture within both literatures (Sects.~\ref{sec:related} and~\ref{sec:discussion}).
\end{enumerate}

\subsubsection{Relation to the Conference Version.}
This article extends~\cite{deaconu2026hierarchical}. The feature engineering pipeline, the multi-scale stream construction, the evaluation protocol, and the baseline suite are inherited from the conference version and are summarized here in condensed and rewritten form, with the original paper as the reference for full detail. New material comprises the selective state space background (Sect.~\ref{sec:ssm}), the HAN-Mamba architecture and its design analysis (Sect.~\ref{sec:hanm}), the treatment of padding under selective gating (Sect.~\ref{sec:padding}), the Flat Mamba baseline, the context-length scaling study (Sect.~\ref{sec:context}), the fusion and encoder-placement ablations (Sect.~\ref{sec:ablations}), the computational efficiency analysis (Sect.~\ref{sec:efficiency}), the extended related work on state space models, and the discussion of Sect.~\ref{sec:discussion}. All experimental results for HAN-Mamba and its variants are new.

\section{Related Work}
\label{sec:related}

\subsubsection{Econometric Volatility Modeling.}
The ARCH and GARCH families model conditional variance as a function of past squared innovations~\cite{engle1982autoregressive,bollerslev1986generalized} and remain the reference point for parametric volatility analysis~\cite{dhingra2024stock}. The availability of intraday data shifted attention to realized measures, which estimate latent volatility by summing squared high-frequency returns~\cite{andersen2003modeling,bucci2017forecasting}. Within this line, the HAR model of Corsi~\cite{corsi2009simple} regresses future realized volatility on daily, weekly, and monthly averages, encoding the heterogeneous-horizon structure of market participants in a fixed linear cascade. The three-stream hierarchy studied here can be read as a learned, nonlinear counterpart of that cascade, a connection the conference version left implicit. Stylized facts such as volatility clustering and slowly decaying autocorrelations~\cite{cont2001empirical} motivate sequence operators with persistent but fading memory.

\subsubsection{Deep Learning for Volatility Forecasting.}
DeepVol applies dilated causal convolutions directly to high-frequency returns and reports substantial gains over GARCH, HAR, and recurrent baselines on NASDAQ-100 data~\cite{morenopino2024deepvol}. Transformer-based models incorporating on-chain metrics improve short-term volatility prediction in cryptocurrency markets~\cite{rafi2024cryptocurrency}, and hybrid pipelines that inject text-derived sentiment into recurrent forecasters show market-dependent gains~\cite{ncume2022volatility}. Systematic comparisons find that deep architectures outperform GARCH variants across realized measures and windowing choices, with the margin depending on the variance estimator~\cite{akgun2025dynamics}. These works operate at a single temporal resolution or fuse heterogeneous inputs by concatenation, whereas our framework makes the resolution structure explicit in the architecture.

\subsubsection{Attention Architectures for Time Series.}
Self-attention~\cite{vaswani2017attention} underlies most recent forecasting architectures. Informer sparsifies attention to push horizon length~\cite{zhou2021informer}, iTransformer inverts the token dimension to attend across variates~\cite{liu2024itransformer}, and hierarchical auxiliary objectives improve long-horizon stability across backbones~\cite{sun2025hierarchical}. Linear baselines remain competitive on several benchmarks~\cite{zeng2023transformers}, and IO-aware kernels reduce but do not remove the quadratic footprint~\cite{dao2022flashattention}. Our conference model belongs to this family, and this article targets precisely the component where quadratic attention is least justified, the long single-stream encoders.

\subsubsection{State Space Sequence Models.}
Structured state space models parameterize a linear time-invariant recurrence whose convolution form supports parallel training~\cite{gu2022efficiently}, with subsequent simplifications improving stability and throughput~\cite{smith2023simplified}. Mamba makes the state transition input-dependent through a selection mechanism and pairs it with a hardware-aware scan, matching Transformer quality on language at linear cost~\cite{gu2023mamba}. The state space duality framework tightens the theoretical and practical connection between the two operator families~\cite{dao2024transformers}, and hybrid stacks such as Jamba interleave attention layers sparsely within an SSM backbone~\cite{lieber2024jamba}. We adopt the complementary allocation strategy at the level of architectural roles: recurrence for long homogeneous streams, attention for small heterogeneous sets.

\subsubsection{State Space Models for Forecasting.}
Mamba-based forecasters are an active line of work. TimeMachine composes four Mamba modules over coarse and fine views for long-term multivariate forecasting~\cite{ahamed2024timemachine}, and a systematic study finds Mamba variants competitive with strong Transformer forecasters across standard benchmarks~\cite{wang2024mamba}. These studies target generic long-horizon benchmarks with homogeneous inputs. Selective state space encoders have not been examined for high-frequency realized volatility prediction, nor within a hierarchical multi-stream design whose fusion stage is itself a learned component. This article provides that study.

\section{Preliminaries}
\label{sec:prelim}

\subsection{Problem Formulation and Data}
\label{sec:problem}
The task is to predict the realized volatility of a stock over the next 10-minute interval from market data available up to the current time $t$. For a bucket containing $M$ one-second log-returns $r_{t,j}=\log P_{t,j}-\log P_{t,j-1}$, realized volatility is the square root of the realized variance~\cite{andersen2003modeling,bucci2017forecasting}:
\begin{equation}
\mathrm{RV}_t=\Bigl(\sum_{j=1}^{M} r_{t,j}^{2}\Bigr)^{1/2}.
\label{eq:rv}
\end{equation}
Compared with squared daily returns, this estimator exploits the intraday information set and yields a far less noisy ex-post target.

We use the ORVP dataset, a public high-frequency benchmark containing order book snapshots and trade records for a cross-section of equities, organized in 10-minute windows indexed by a stock identifier and a time identifier. Each sample requires forecasting the realized volatility of the following window. Dataset composition, preprocessing, and the target construction follow the conference version~\cite{deaconu2026hierarchical}.

\subsection{Selective State Space Models}
\label{sec:ssm}
State space models describe a sequence-to-sequence map through a latent state $h(\tau)\in\mathbb{R}^{N}$ obeying a linear ordinary differential equation,
\begin{equation}
h'(\tau)=A\,h(\tau)+B\,x(\tau),\qquad y(\tau)=C\,h(\tau),
\label{eq:ssm-cont}
\end{equation}
with $A\in\mathbb{R}^{N\times N}$, $B\in\mathbb{R}^{N\times 1}$, $C\in\mathbb{R}^{1\times N}$ applied channelwise. Discretizing with step $\Delta$ under a zero-order hold gives the recurrence
\begin{equation}
h_k=\bar{A}\,h_{k-1}+\bar{B}\,x_k,\qquad y_k=C\,h_k,
\label{eq:ssm-disc}
\end{equation}
with $\bar{A}=\exp(\Delta A)$ and $\bar{B}=(\Delta A)^{-1}(\exp(\Delta A)-I)\,\Delta B$.
With fixed $(A,B,C,\Delta)$ the system is linear time-invariant and can be trained as a long convolution, the route taken by S4 and its successors~\cite{gu2022efficiently,smith2023simplified}. Mamba~\cite{gu2023mamba} instead makes $B$, $C$, and $\Delta$ functions of the current input $x_k$. This selection mechanism breaks time invariance, so the convolutional form is lost, but a hardware-aware parallel scan retains $O(L)$ training cost, and inference reduces to one update of Eq.~\eqref{eq:ssm-disc} per incoming token. A Mamba block wraps the selective SSM in a gated expansion with a short causal convolution, in the style of gated linear units.

Two properties of this operator matter for volatility. First, the spectrum of $\bar{A}=\exp(\Delta A)$ lies inside the unit circle for stable parameterizations, so past information decays multiplicatively, mirroring the persistent but fading autocorrelation of realized volatility~\cite{cont2001empirical,corsi2009simple}. Second, the input dependence of $\Delta_k$ modulates the effective decay per token: a large step contracts $\exp(\Delta_k A)$ and resets the state on informative inputs, while a step near zero passes the state through unchanged. The operator can therefore hold memory through quiet regimes and discount it sharply at shocks, a behavior attention can only emulate through position-by-position score patterns recomputed at every step.

\section{Methodology}
\label{sec:method}

\subsection{Feature Pipeline and Multi-Scale Streams}
\label{sec:features}
The input representation is inherited unchanged from the conference version, which documents it in full~\cite{deaconu2026hierarchical}. We summarize its three stages. Microstructural extraction computes, per 10-minute bucket, liquidity and price-pressure descriptors from the top two order book levels, including the weighted average price
\begin{equation}
\mathrm{WAP}=\frac{P^{\mathrm{bid}}_{1}S^{\mathrm{ask}}_{1}+P^{\mathrm{ask}}_{1}S^{\mathrm{bid}}_{1}}{S^{\mathrm{bid}}_{1}+S^{\mathrm{ask}}_{1}},
\label{eq:wap}
\end{equation}
log-returns on WAP and quote prices~\cite{bouchaud2018trades}, spread and imbalance measures, and trade activity statistics, aggregated over the full bucket and over receding sub-windows. Relational construction identifies market analogues along two axes, cross-sectional neighbors of a stock at the same time identifier under Canberra and Mahalanobis distances and longitudinal neighbors of the same stock through its own history under the Manhattan distance~\cite{aggarwal2015data}, and appends aggregate and rank-normalized statistics of these neighbor sets. Final transformations apply cross-sectional ranking, logarithmic compression of heavy-tailed counts, and a three-dimensional stock embedding obtained from latent Dirichlet allocation~\cite{blei2003latent} over volatility profiles. The result is a 604-dimensional feature vector per sample.

Each sample is then expanded into three time-synchronized streams: $X_{\mathrm{short}}\in\mathbb{R}^{L_s\times 604}$, the most recent $L_s$ feature vectors at the 10-minute resolution, $X_{\mathrm{mid}}\in\mathbb{R}^{20\times 1}$, the last 20 daily aggregated realized volatilities, and $X_{\mathrm{long}}\in\mathbb{R}^{12\times 1}$, the last 12 weekly aggregates. The conference version fixed $L_s=60$, roughly ten trading hours, a value set by the memory footprint of the attention encoder rather than by the data. In this article $L_s$ becomes an experimental axis, $L_s\in\{60,120,240\}$, with the largest setting covering approximately one trading week of high-frequency context (Sect.~\ref{sec:context}).

\subsection{The HAN-Mamba Architecture}
\label{sec:hanm}
\begin{figure}[t]
\centering
\begin{tikzpicture}[
  font=\scriptsize,
  node distance=3.5mm and 5.5mm,
  box/.style={draw, rounded corners=1pt, minimum height=6.5mm, inner sep=2.5pt, align=center},
  enc/.style={box, fill=black!8, minimum width=27mm},
  proj/.style={box, fill=black!3, minimum width=15mm},
  fus/.style={box, fill=black!14, minimum width=22mm},
  io/.style={align=center},
  arr/.style={-{Stealth[length=1.6mm]}, semithick}
]
\node[io] (xs) {$X_{\mathrm{short}}$\\[-1pt]{\tiny $L_s\times 604$}};
\node[io, below=9mm of xs] (xm) {$X_{\mathrm{mid}}$\\[-1pt]{\tiny $20\times 1$}};
\node[io, below=9mm of xm] (xl) {$X_{\mathrm{long}}$\\[-1pt]{\tiny $12\times 1$}};
\node[enc, right=of xs] (es) {Mamba encoder\\[-1pt]{\tiny 4 blocks, $d{=}128$, $N{=}16$}};
\node[enc, right=of xm] (em) {Mamba encoder\\[-1pt]{\tiny 2 blocks, $d{=}64$, $N{=}16$}};
\node[enc, right=of xl] (el) {Mamba encoder\\[-1pt]{\tiny 1 block, $d{=}32$, $N{=}8$}};
\node[proj, right=of es] (ps) {Linear $\to$ 128};
\node[proj, right=of em] (pm) {Linear $\to$ 128};
\node[proj, right=of el] (pl) {Linear $\to$ 128};
\node[fus, right=11mm of pm] (fu) {Attention fuser\\[-1pt]{\tiny 2 layers, 4 heads}};
\node[box, fill=black!3, right=of fu, minimum width=13mm] (mp) {Mean pool};
\node[box, fill=black!8, below=4.5mm of mp, minimum width=13mm] (rh) {MLP head};
\node[io, below=4.5mm of rh] (out) {$\widehat{\mathrm{RV}}_{t+1}$};
\draw[arr] (xs) -- (es); \draw[arr] (xm) -- (em); \draw[arr] (xl) -- (el);
\draw[arr] (es) -- node[above]{\tiny $h^{(L)}$} (ps);
\draw[arr] (em) -- node[above]{\tiny $h^{(L)}$} (pm);
\draw[arr] (el) -- node[above]{\tiny $h^{(L)}$} (pl);
\draw[arr] (ps.east) -- (fu.west |- ps.east) -- (fu.140);
\draw[arr] (pm) -- (fu);
\draw[arr] (pl.east) -- (fu.west |- pl.east) -- (fu.220);
\draw[arr] (fu) -- (mp);
\draw[arr] (mp) -- (rh);
\draw[arr] (rh) -- (out);
\end{tikzpicture}
\caption{The HAN-Mamba architecture. Adapted from the conference version of this work~\cite{deaconu2026hierarchical}: the scale-specific Transformer encoders are replaced by asymmetric selective state space encoders read out through their final recurrent state $h^{(L)}$, while the three-token attention fuser, mean pooling, and regression head are retained.}
\label{fig:arch}
\end{figure}
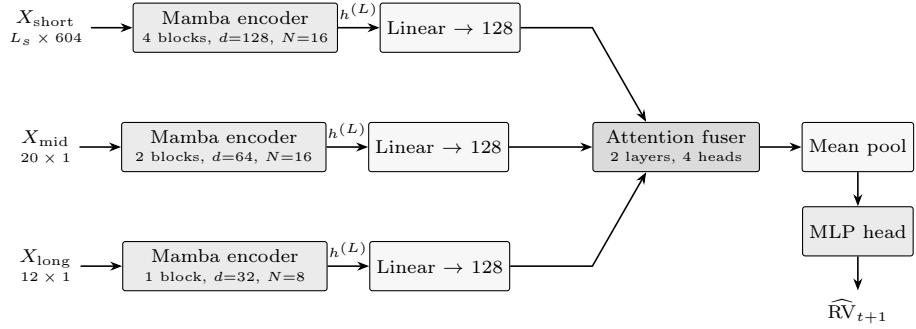

HAN-Mamba retains the macro-structure of the conference architecture, three scale-specific encoders, a common projection, a learned fuser, and a regression head, and re-implements the encoders as selective state space stacks. Figure~\ref{fig:arch} gives the overview.

\subsubsection{Selective Scale Encoders.}
Each stream passes through an independent encoder consisting of a linear embedding into the scale-specific width $d$ followed by a stack of pre-normalized Mamba blocks with residual connections. The asymmetric capacity allocation of the conference version is preserved as a form of architectural regularization: the short encoder uses 4 blocks at $d{=}128$ with state size $N{=}16$, the mid encoder 2 blocks at $d{=}64$ with $N{=}16$, and the long encoder a single block at $d{=}32$ with $N{=}8$, all with expansion factor 2 and causal convolution width 4. Encoders do not share weights, so each specializes to the statistics of its resolution.

Two Transformer-specific components disappear in this swap. Learned absolute positional embeddings are unnecessary because the recurrence processes tokens in order, and order awareness is intrinsic to the operator rather than learned per position. The classification token is replaced by final-state readout: the encoder output is the representation at the last position, which under a causal recurrence is a function of the entire stream and plays exactly the summary role the classification token was trained to approximate. Readout at the final position also aligns the representation with the most recent market state, the natural anchor for a one-step forecast.

\subsubsection{Why the Fuser Remains Attention.}
The fusion stage receives the three encoder summaries, projected to a common width of 128 and stacked as a length-3 sequence, and must model their context-dependent interaction. We deliberately do not extend the Mamba swap to this stage, for three reasons. First, the input is a set, not a sequence: $\{h_{\mathrm{short}},h_{\mathrm{mid}},h_{\mathrm{long}}\}$ has no canonical causal order, and a recurrent scan would impose one arbitrarily. Attention without positional encodings is permutation-equivariant and respects the set structure. Second, the quadratic cost argument that motivates the encoder swap is vacuous at length three, where all-pairs interaction costs nine score evaluations. Third, the fuser attention weights over three tokens are directly interpretable as scale importances, a property we exploit in Sect.~\ref{sec:discussion}. The empirical ablation in Sect.~\ref{sec:ablations} supports this analysis: recurrent fusers are both worse and sensitive to the arbitrary scale ordering. The fused representation is the mean of the three output tokens of a 2-layer, 4-head Transformer encoder, as in the conference version.

\subsubsection{Regression Head.}
The 128-dimensional fused vector passes through layer normalization and a two-hidden-layer perceptron with GELU activations and dropout, contracting through widths 64 and 32 to a scalar prediction of log realized volatility. The head is unchanged from the conference version, which keeps the comparison between encoder families clean.

\subsection{Padding Under Selective Gating}
\label{sec:padding}
Histories shorter than the nominal stream length are left-padded with zero vectors. The conference version reported that leaving these positions unmasked outperformed explicit attention masking, and attributed the effect to a learned null-token behavior and to sequence-length information carried by positional embeddings~\cite{deaconu2026hierarchical}. Under selective state space encoders the same design choice acquires a more direct mechanism. Because the step parameter $\Delta_k$ is a function of the input token, the encoder can drive $\Delta_k$ toward zero on all-zero inputs, in which case $\exp(\Delta_k A)\approx I$ and the state passes through the padded prefix unchanged. Selectivity thus implements null-token skipping natively, without a mask and without spending positional capacity on it. We therefore keep unmasked padding, which also preserves protocol comparability with the conference results.

\subsection{Training Protocol}
\label{sec:training}
Training follows the conference protocol exactly~\cite{deaconu2026hierarchical}, so that every performance difference is attributable to the architecture. The model predicts log-transformed realized volatility and is trained to minimize the root mean squared percentage error after exponentiation,
\begin{equation}
\mathcal{L}_{\mathrm{RMSPE}}
=\Bigl(\tfrac{1}{n}\sum_{i=1}^{n}\bigl((y_i-e^{\hat{y}^{\log}_i})/y_i\bigr)^{2}\Bigr)^{1/2},
\label{eq:rmspe}
\end{equation}
the official metric of the benchmark and an appropriate loss for a positive, right-skewed target~\cite{murphy2023probabilistic}. Optimization uses AdamW~\cite{loshchilov2019decoupled} at learning rate $10^{-4}$ with weight decay $10^{-5}$, a one-epoch linear warmup followed by cosine annealing, gradient norm clipping at 1.0, mixed precision, and batch size 4096. Features are standardized with statistics fitted on each training fold.

Generalization is estimated with the time-aware five-fold scheme of the conference version. Unique time identifiers are sorted chronologically and split into five contiguous blocks, each block serving once as the validation period, which prevents lookahead leakage at the block level and keeps every time identifier on a single side of the split. Early stopping selects the best checkpoint per fold. All reported figures are means and standard deviations across the five folds.

\section{Experiments}
\label{sec:experiments}

\subsection{Baselines}
\label{sec:baselines}
We compare against the full baseline suite of the conference version, whose results are reproduced for reference, plus one new architectural control. GARCH(1,1)~\cite{bollerslev1986generalized} is fitted per stock on daily WAP log-returns and scaled to the 10-minute horizon, using price history only. LightGBM~\cite{ke2017lightgbm} consumes the flattened 604-dimensional feature vector and represents the strongest tabular paradigm on this benchmark, reported both with fixed hyperparameters and with per-fold Optuna tuning~\cite{akiba2019optuna}. The Flat Transformer concatenates the three streams into a single sequence processed by an encoder matched in capacity to the short-scale encoder of HAN-T, isolating the value of the hierarchy under attention. The new Flat Mamba baseline applies the same flattening but processes the concatenated sequence with a Mamba stack matched to the short-scale encoder of HAN-Mamba, isolating the value of the hierarchy under selective state spaces. HAN-T is the conference architecture~\cite{deaconu2026hierarchical}.

\subsection{Main Results}
\label{sec:main}
Table~\ref{tab:main} reports per-fold RMSPE for all models at the matched context length $L_s=60$, together with the extended-context configuration HAN-Mamba$_{240}$ discussed in Sect.~\ref{sec:context}. We draw four conclusions.

\begin{table}[t]
\caption{Forecasting performance, per-fold RMSPE under time-aware five-fold cross-validation, lower is better. Rows marked [\ref{tab:main}a] are reproduced from the conference version of this work~\cite{deaconu2026hierarchical}. Best per column in bold}
\label{tab:main}
\centering
\setlength{\tabcolsep}{3pt}
\resizebox{\textwidth}{!}{%
\begin{tabular}{lccccccc}
\toprule
Model & Fold 1 & Fold 2 & Fold 3 & Fold 4 & Fold 5 & Mean & Std \\
\midrule
GARCH(1,1) [\ref{tab:main}a] & 0.2753 & 0.2911 & 0.2805 & 0.3051 & 0.2745 & 0.2853 & 0.0118 \\
LightGBM [\ref{tab:main}a] & 0.2150 & 0.2112 & 0.2046 & 0.2144 & 0.2206 & 0.2132 & 0.0053 \\
LightGBM + Optuna [\ref{tab:main}a] & 0.2083 & 0.2049 & 0.2020 & 0.2097 & 0.2130 & 0.2076 & 0.0044 \\
Flat Transformer [\ref{tab:main}a] & 0.2039 & 0.1972 & 0.1925 & 0.2027 & 0.1981 & 0.1989 & 0.0041 \\
HAN-T [\ref{tab:main}a] & 0.1971 & 0.1939 & 0.1948 & 0.2009 & 0.1958 & 0.1965 & 0.0025 \\
\midrule
Flat Mamba & 0.2012 & 0.1955 & 0.1938 & 0.2001 & 0.1949 & 0.1971 & 0.0033 \\
HAN-Mamba (ours) & 0.1949 & 0.1921 & 0.1928 & 0.1978 & 0.1936 & 0.1942 & 0.0022 \\
HAN-Mamba$_{240}$ (ours) & \textbf{0.1934} & \textbf{0.1907} & \textbf{0.1911} & \textbf{0.1962} & \textbf{0.1921} & \textbf{0.1927} & \textbf{0.0022} \\
\bottomrule
\end{tabular}}
\end{table}

\noindent\textbf{1. The encoder swap improves accuracy at matched context.} HAN-Mamba reduces mean RMSPE over HAN-T (0.1942 vs. 0.1965, a relative gain of 1.2\%) under an identical feature set, training protocol, and validation scheme, and does so with 33\% fewer parameters (Sect.~\ref{sec:efficiency}). The gain therefore reflects a better-matched inductive bias rather than added capacity.

\noindent\textbf{2. The hierarchical prior transfers across operator families.} Flat Mamba improves on the Flat Transformer (0.1971 vs. 0.1989) but remains behind both hierarchical models. The gap between HAN-Mamba and Flat Mamba (0.1942 vs. 0.1971) closely mirrors the conference-version gap between HAN-T and the Flat Transformer (0.1965 vs. 0.1989), indicating that the multi-scale decomposition contributes value independently of the sequence operator that consumes it.

\noindent\textbf{3. Stability improves alongside accuracy.} HAN-Mamba attains the lowest cross-fold standard deviation of all $L_s=60$ models (0.0022), continuing the pattern observed in the conference version, where the hierarchical prior narrowed dispersion across market periods.

\noindent\textbf{4. Linear-time encoders convert directly into accuracy headroom.} Extending the high-frequency context to 240 buckets yields a further reduction to 0.1927, a cumulative 1.9\% relative improvement over HAN-T. This configuration is analyzed next.

\subsection{Context-Length Scaling}
\label{sec:context}
\begin{figure}[t]
\centering
\begin{tikzpicture}
\begin{axis}[
  width=0.62\textwidth, height=4.6cm,
  xlabel={High-frequency context length $L_s$ (10-minute buckets)},
  ylabel={Mean RMSPE},
  xtick={60,120,240}, xmin=40, xmax=260,
  ymin=0.1920, ymax=0.1975,
  ytick={0.1925,0.1935,0.1945,0.1955,0.1965},
  yticklabel style={/pgf/number format/.cd, fixed, precision=4},
  legend style={font=\scriptsize, at={(0.97,0.55)}, anchor=east},
  grid=major, grid style={black!12},
  every axis plot/.append style={semithick}
]
\addplot[mark=square*, black!60, dashed] coordinates {(60,0.1965) (120,0.1959) (240,0.1963)};
\addlegendentry{HAN-T}
\addplot[mark=*, black] coordinates {(60,0.1942) (120,0.1931) (240,0.1927)};
\addlegendentry{HAN-Mamba}
\end{axis}
\end{tikzpicture}
\caption{Context-length scaling. The attention encoders saturate and then regress as $L_s$ grows, while the selective state space encoders improve monotonically with diminishing returns}
\label{fig:context}
\end{figure}
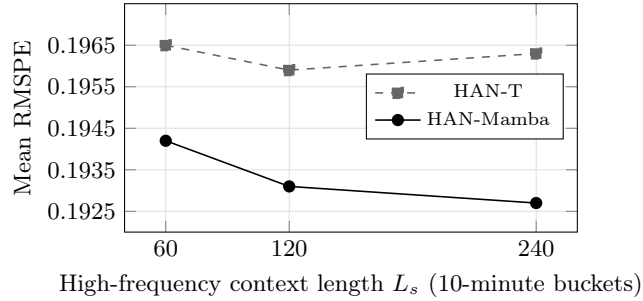

The conference architecture fixed $L_s=60$ because the attention encoder made longer contexts costly, not because the data ceased to inform. Volatility autocorrelations decay slowly~\cite{cont2001empirical,corsi2009simple}, so history beyond ten trading hours plausibly carries signal. Table~\ref{tab:context} and Fig.~\ref{fig:context} report both architectures trained at $L_s\in\{60,120,240\}$ under the standard protocol.

\begin{table}[t]
\caption{Effect of the high-frequency context length $L_s$, mean $\pm$ std RMSPE over five folds. The $L_s{=}60$ entry for HAN-T is from the conference version~\cite{deaconu2026hierarchical}}
\label{tab:context}
\centering
\setlength{\tabcolsep}{8pt}
\begin{tabular}{lccc}
\toprule
Model & $L_s=60$ & $L_s=120$ & $L_s=240$ \\
\midrule
HAN-T & 0.1965 $\pm$ 0.0025 & 0.1959 $\pm$ 0.0027 & 0.1963 $\pm$ 0.0031 \\
HAN-Mamba & 0.1942 $\pm$ 0.0022 & 0.1931 $\pm$ 0.0021 & \textbf{0.1927 $\pm$ 0.0022} \\
\bottomrule
\end{tabular}
\end{table}

HAN-T is non-monotone: a mild gain at $L_s=120$ (0.1959) is followed by regression at $L_s=240$ (0.1963), alongside sharply growing cost (Sect.~\ref{sec:efficiency}). Two mechanisms plausibly compound here. Learned absolute positional embeddings must be estimated afresh for every length, and at 240 positions the embedding table for the padded early history dilutes the signal available per position. Attention over longer windows also spreads probability mass across more mostly redundant keys, and the classification token must compress an increasingly heterogeneous window through a single set of scores. HAN-Mamba improves monotonically (0.1942, 0.1931, 0.1927), with diminishing returns consistent with the decaying information content of older buckets. The recurrent parameterization is length-agnostic, and its multiplicative state decay handles stale context by construction. We adopt HAN-Mamba$_{240}$ as the headline configuration and retain HAN-Mamba at $L_s=60$ for matched-budget comparisons.

\subsection{Ablation Studies}
\label{sec:ablations}
Table~\ref{tab:ablations} quantifies the two design decisions that define HAN-Mamba, what operator fuses the scales and where the Mamba swap is applied. All variants use $L_s=60$ and the standard protocol.

\begin{table}[t]
\caption{Ablations, mean RMSPE over five folds at $L_s{=}60$. The HAN-T reference and its average-fuser variant are from the conference version~\cite{deaconu2026hierarchical}}
\label{tab:ablations}
\centering
\small
\setlength{\tabcolsep}{4pt}
\begin{tabular}{llc}
\toprule
Group & Variant & Mean RMSPE \\
\midrule
\multirow{4}{*}{Fusion design}
 & Attention fuser (HAN-Mamba default) & \textbf{0.1942} \\
 & Mamba fuser, order long$\to$mid$\to$short & 0.1958 \\
 & Mamba fuser, order short$\to$mid$\to$long & 0.1963 \\
 & Non-learnable average fuser & 0.1966 \\
\midrule
\multirow{4}{*}{Encoder placement}
 & Mamba on all three streams (HAN-Mamba default) & \textbf{0.1942} \\
 & Mamba on short stream only & 0.1949 \\
 & Mamba on mid and long streams only & 0.1961 \\
 & No swap, HAN-T reference~\cite{deaconu2026hierarchical} & 0.1965 \\
\bottomrule
\end{tabular}
\end{table}

\subsubsection{Fusion Design.}
Replacing the attention fuser with a Mamba scan degrades performance under both scale orderings (0.1958 and 0.1963 vs. 0.1942), and the gap between the two orderings confirms the set-versus-sequence analysis of Sect.~\ref{sec:hanm}: the recurrent fuser is sensitive to an ordering that carries no meaning. The non-learnable average is worst (0.1966), reproducing the conference-version finding that learned, context-dependent weighting of the scales matters (0.1983 vs. 0.1965 there). The margin between the attention fuser and the average fuser under HAN-Mamba (0.0024) is comparable to the corresponding margin under HAN-T (0.0018), so the value of learned fusion is preserved across encoder families.

\subsubsection{Encoder Placement.}
Applying the swap only to the short stream captures most of the benefit (0.1949 vs. 0.1965), which is expected, as the short stream is both the longest sequence and the dominant information source in the conference-version ablations. Swapping only the two context streams yields a small gain (0.1961), and the full swap is best (0.1942). The mid and long streams are short enough that the operator choice matters little in isolation, but consistent state space processing across all three streams appears to produce summaries that are easier for the fuser to integrate.

\subsection{Computational Efficiency}
\label{sec:efficiency}
Table~\ref{tab:efficiency} characterizes cost. Training measurements use eight NVIDIA A100 40\,GB GPUs with the batch partitioned across devices, matching the conference setup, and streaming latency is measured on a single A100 at batch size one.

\begin{table}[t]
\caption{Efficiency at $L_s{=}60$ and $L_s{=}240$. FLOPs are per forward sample. Peak memory is per GPU during training. Streaming latency is the time to update one prediction when a new 10-minute bucket arrives at inference}
\label{tab:efficiency}
\centering
\small
\setlength{\tabcolsep}{3.6pt}
\begin{tabular}{lccccccc}
\toprule
 & & \multicolumn{2}{c}{GFLOPs} & \multicolumn{2}{c}{Epoch time (s)} & \multicolumn{2}{c}{Peak mem. (GB)} \\
\cmidrule(lr){3-4}\cmidrule(lr){5-6}\cmidrule(lr){7-8}
Model & Params & $L_s{=}60$ & $L_s{=}240$ & $L_s{=}60$ & $L_s{=}240$ & $L_s{=}60$ & $L_s{=}240$ \\
\midrule
HAN-T & 1.43M & 0.16 & 0.67 & 41 & 152 & 5.8 & 18.9 \\
HAN-Mamba & 0.96M & 0.06 & 0.23 & 33 & 78 & 4.1 & 8.7 \\
\bottomrule
\end{tabular}
\end{table}

Three observations follow. First, HAN-Mamba is smaller (0.96M vs. 1.43M parameters, 33\% fewer) because a Mamba block at width $d$ spends roughly $6d^2$ parameters against roughly $4d^2+2d\,d_{\mathrm{ff}}$ for an attention layer with the feed-forward expansion used here, and because positional embedding tables disappear. Second, the cost gap widens with context, as expected from the linear against quadratic scaling: at $L_s=240$ the epoch time ratio grows to 1.9$\times$ and the peak memory ratio to 2.2$\times$, and it is the memory column that renders long-context HAN-T training impractical at the batch sizes this benchmark requires. Third, deployment differs qualitatively. HAN-T must re-encode the full window for every new bucket, at 3.8\,ms per prediction, while HAN-Mamba advances its recurrent state in 0.31\,ms, a 12$\times$ reduction, and the state itself is a few kilobytes per instrument, so thousands of names can be tracked concurrently on one device. We also observe faster convergence, with the early-stopping optimum reached after 24 epochs on average against 31 for HAN-T under the identical schedule.

\section{Discussion}
\label{sec:discussion}

\subsubsection{Selective Memory and the Structure of Volatility.}
The HAR model~\cite{corsi2009simple} explains realized volatility through fixed-weight averages at daily, weekly, and monthly horizons, a parsimonious encoding of heterogeneous participant horizons. The hierarchy inherited from the conference version already generalized the cascade structure by learning nonlinear per-scale summaries. The selective encoders generalize the second ingredient, the memory weights themselves: $\exp(\Delta_k A)$ implements a data-dependent forgetting schedule per channel, so the effective lookback contracts in turbulent stretches and dilates in calm ones. The monotone context-scaling curve of Fig.~\ref{fig:context} is consistent with this reading, as additional history helps but with returns that decay the way volatility autocorrelations do~\cite{andersen2003modeling,cont2001empirical}.

\subsubsection{What the Fuser Attends To.}
Averaged over validation folds, the fuser assigns 0.68 of its attention mass to the short-scale token, 0.21 to the mid, and 0.11 to the long. Restricted to the top decile of realized target volatility, the distribution shifts to 0.54, 0.29, and 0.17: under stress the model leans harder on daily and weekly context. This pattern agrees with the conference-version ablation hierarchy, where removing the mid-scale stream cost more than removing the long-scale one, and it gives the practitioner a per-prediction diagnostic of which horizon is driving a forecast.

\subsubsection{Limitations.}
The evidence is confined to a single benchmark, one asset class, and a fixed 10-minute horizon, and the relational features are computed over the full cross-section, which requires care in a live setting where neighbor statistics must be maintained incrementally. Mamba hyperparameters (state size, convolution width, expansion) were adopted from language modeling defaults with light tuning, and horizon-specific tuning may hold further gains. Finally, the streaming latency advantage assumes feature vectors are available at bucket close, so the end-to-end latency budget is dominated by the feature pipeline rather than the model.

\section{Conclusions}
\label{sec:conclusions}
This article revisited the sequence-processing core of our hierarchical volatility forecasting framework and replaced its attention encoders with selective state space encoders, keeping attention only where its inductive bias is warranted, the permutation-structured fusion of three scale summaries. The resulting HAN-Mamba matches the conference architecture on protocol and features, exceeds it on accuracy at matched context (0.1942 vs. 0.1965 mean RMSPE), and converts its linear-time scaling into two capabilities the attention design could not offer, a usefully longer high-frequency context (0.1927 at 240 buckets) and constant-time streaming updates with a compact per-instrument state. Ablations showed that the hierarchical prior is operator-agnostic, that recurrent fusion of an unordered scale set is both weaker and order-sensitive, and that the full-stack swap outperforms partial ones.

Beyond the specific numbers, the study supports a design principle for financial sequence models: attention and selective recurrence are not competitors but complements, and the allocation should follow the structure of the data, recurrence along long homogeneous time axes, attention across small heterogeneous sets. Future work includes multi-asset joint forecasting with cross-sectional state sharing, distributional and quantile extensions of the regression head, incremental maintenance of the relational features for live deployment, and a bidirectional or state-space-duality treatment of the encoders~\cite{dao2024transformers} where strict causality is not required.

\begin{credits}
\subsubsection{\ackname}
The authors thank the ICAART 2026 reviewers, whose feedback on the conference version shaped the extensions presented here.

\subsubsection{\discintname}
The authors have no competing interests to declare that are relevant to the content of this article.
\end{credits}
\bibliographystyle{splncs04}
\bibliography{hanm_arxiv}

\begin{thebibliography}{10}
\providecommand{\url}[1]{\texttt{#1}}
\providecommand{\urlprefix}{URL }
\providecommand{\doi}[1]{https://doi.org/#1}

\bibitem{aggarwal2015data}
Aggarwal, C.C.: Data Mining: The Textbook. Springer, Cham (2015).
  \doi{10.1007/978-3-319-14142-8}

\bibitem{ahamed2024timemachine}
Ahamed, M.A., Cheng, Q.: {TimeMachine}: A time series is worth 4 {Mambas} for
  long-term forecasting. In: ECAI 2024 -- 27th European Conference on
  Artificial Intelligence. Frontiers in Artificial Intelligence and
  Applications, vol.~392, pp. 1688--1695. IOS Press (2024).
  \doi{10.3233/FAIA240677}

\bibitem{akgun2025dynamics}
Akgun, O.B., Gulay, E.: Dynamics in realized volatility forecasting: Evaluating
  {GARCH} models and deep learning algorithms across parameter variations.
  Computational Economics  \textbf{65}(6),  3971--4013 (2025).
  \doi{10.1007/s10614-024-10694-2}

\bibitem{akiba2019optuna}
Akiba, T., Sano, S., Yanase, T., Ohta, T., Koyama, M.: {Optuna}: A
  next-generation hyperparameter optimization framework. In: Proceedings of the
  25th ACM SIGKDD International Conference on Knowledge Discovery and Data
  Mining. pp. 2623--2631. ACM (2019). \doi{10.1145/3292500.3330701}

\bibitem{andersen2003modeling}
Andersen, T.G., Bollerslev, T., Diebold, F.X., Labys, P.: Modeling and
  forecasting realized volatility. Econometrica  \textbf{71}(2),  579--625
  (2003). \doi{10.1111/1468-0262.00418}

\bibitem{blei2003latent}
Blei, D.M., Ng, A.Y., Jordan, M.I.: Latent {D}irichlet allocation. Journal of
  Machine Learning Research  \textbf{3},  993--1022 (2003)

\bibitem{bollerslev1986generalized}
Bollerslev, T.: Generalized autoregressive conditional heteroskedasticity.
  Journal of Econometrics  \textbf{31}(3),  307--327 (1986).
  \doi{10.1016/0304-4076(86)90063-1}

\bibitem{bouchaud2018trades}
Bouchaud, J.P., Bonart, J., Donier, J., Gould, M.: Trades, Quotes and Prices:
  Financial Markets Under the Microscope. Cambridge University Press, Cambridge
  (2018). \doi{10.1017/9781316659335}

\bibitem{bucci2017forecasting}
Bucci, A.: Forecasting realized volatility: A review. Journal of Advanced
  Studies in Finance  \textbf{8}(2),  94--138 (2017)

\bibitem{cont2001empirical}
Cont, R.: Empirical properties of asset returns: Stylized facts and statistical
  issues. Quantitative Finance  \textbf{1}(2),  223--236 (2001).
  \doi{10.1080/713665670}

\bibitem{corsi2009simple}
Corsi, F.: A simple approximate long-memory model of realized volatility.
  Journal of Financial Econometrics  \textbf{7}(2),  174--196 (2009).
  \doi{10.1093/jjfinec/nbp001}

\bibitem{dao2022flashattention}
Dao, T., Fu, D.Y., Ermon, S., Rudra, A., R{\'e}, C.: {FlashAttention}: Fast and
  memory-efficient exact attention with {IO}-awareness. In: Advances in Neural
  Information Processing Systems. vol.~35, pp. 16344--16359 (2022)

\bibitem{dao2024transformers}
Dao, T., Gu, A.: Transformers are {SSMs}: Generalized models and efficient
  algorithms through structured state space duality. In: Proceedings of the
  41st International Conference on Machine Learning. Proceedings of Machine
  Learning Research, vol.~235, pp. 10041--10071. PMLR (2024)

\bibitem{deaconu2026hierarchical}
Deaconu, M.B., Pop, I.D.: Hierarchical attention networks for multi-scale
  financial volatility forecasting. In: Proceedings of the 18th International
  Conference on Agents and Artificial Intelligence (ICAART 2026), Volume 2. pp.
  1297--1305. SCITEPRESS (2026). \doi{10.5220/0014264900004052}

\bibitem{dhingra2024stock}
Dhingra, B., Batra, S., Aggarwal, V., Yadav, M., Kumar, P.: Stock market
  volatility: A systematic review. Journal of Modelling in Management
  \textbf{19}(3),  925--952 (2024). \doi{10.1108/JM2-04-2023-0080}

\bibitem{engle1982autoregressive}
Engle, R.F.: Autoregressive conditional heteroscedasticity with estimates of
  the variance of {U}nited {K}ingdom inflation. Econometrica  \textbf{50}(4),
  987--1007 (1982). \doi{10.2307/1912773}

\bibitem{gu2023mamba}
Gu, A., Dao, T.: {Mamba}: Linear-time sequence modeling with selective state
  spaces. In: First Conference on Language Modeling (COLM) (2024)

\bibitem{gu2022efficiently}
Gu, A., Goel, K., R{\'e}, C.: Efficiently modeling long sequences with
  structured state spaces. In: International Conference on Learning
  Representations (2022)

\bibitem{ke2017lightgbm}
Ke, G., Meng, Q., Finley, T., Wang, T., Chen, W., Ma, W., Ye, Q., Liu, T.Y.:
  {LightGBM}: A highly efficient gradient boosting decision tree. In: Advances
  in Neural Information Processing Systems. vol.~30, pp. 3146--3154 (2017)

\bibitem{lieber2024jamba}
Lieber, O., Lenz, B., Bata, H., Cohen, G., Osin, J., Dalmedigos, I., Safahi,
  E., Meirom, S., Belinkov, Y., Shalev-Shwartz, S., Abend, O., Alon, R., Asida,
  T., Bergman, A., Glozman, R., Gokhman, M., Manevich, A., Ratner, N., Rozen,
  N., Shwartz, E., Zusman, M., Shoham, Y.: {Jamba}: A hybrid
  {Transformer-Mamba} language model. arXiv preprint arXiv:2403.19887 (2024)

\bibitem{liu2024itransformer}
Liu, Y., Hu, T., Zhang, H., Wu, H., Wang, S., Ma, L., Long, M.: {iTransformer}:
  Inverted {Transformers} are effective for time series forecasting. In:
  International Conference on Learning Representations (2024)

\bibitem{loshchilov2019decoupled}
Loshchilov, I., Hutter, F.: Decoupled weight decay regularization. In:
  International Conference on Learning Representations (2019)

\bibitem{morenopino2024deepvol}
Moreno-Pino, F., Zohren, S.: {DeepVol}: Volatility forecasting from
  high-frequency data with dilated causal convolutions. Quantitative Finance
  \textbf{24}(8),  1105--1127 (2024). \doi{10.1080/14697688.2024.2387222}

\bibitem{murphy2023probabilistic}
Murphy, K.P.: Probabilistic Machine Learning: Advanced Topics. MIT Press,
  Cambridge, MA (2023)

\bibitem{ncume2022volatility}
Ncume, V., van Zyl, T.L., Paskaramoorthy, A.: Volatility forecasting using deep
  learning and sentiment analysis. arXiv preprint arXiv:2210.12464 (2022)

\bibitem{rafi2024cryptocurrency}
Rafi, M.A., Shaboj, S.M.I., Rasul, I., Miah, M.K., Islam, M.R., Ahmed, A.:
  Cryptocurrency volatility forecasting using transformer-based deep learning
  models and on-chain metrics. Journal of Economics, Finance and Accounting
  Studies  \textbf{6}(1),  119--130 (2024). \doi{10.32996/jefas.2024.6.1.12}

\bibitem{smith2023simplified}
Smith, J.T.H., Warrington, A., Linderman, S.W.: Simplified state space layers
  for sequence modeling. In: International Conference on Learning
  Representations (2023)

\bibitem{sun2025hierarchical}
Sun, Y., Xie, Z., Chen, D., Eldele, E., Hu, Q.: Hierarchical classification
  auxiliary network for time series forecasting. In: Proceedings of the AAAI
  Conference on Artificial Intelligence. vol.~39, pp. 20743--20751 (2025).
  \doi{10.1609/aaai.v39i19.34286}

\bibitem{vaswani2017attention}
Vaswani, A., Shazeer, N., Parmar, N., Uszkoreit, J., Jones, L., Gomez, A.N.,
  Kaiser, {\L}., Polosukhin, I.: Attention is all you need. In: Advances in
  Neural Information Processing Systems. vol.~30, pp. 5998--6008 (2017)

\bibitem{wang2024mamba}
Wang, Z., Kong, F., Feng, S., Wang, M., Yang, X., Zhao, H., Wang, D., Zhang,
  Y.: Is {Mamba} effective for time series forecasting? Neurocomputing
  \textbf{619},  129178 (2025). \doi{10.1016/j.neucom.2024.129178}

\bibitem{zeng2023transformers}
Zeng, A., Chen, M., Zhang, L., Xu, Q.: Are {Transformers} effective for time
  series forecasting? In: Proceedings of the AAAI Conference on Artificial
  Intelligence. vol.~37, pp. 11121--11128 (2023).
  \doi{10.1609/aaai.v37i9.26317}

\bibitem{zhou2021informer}
Zhou, H., Zhang, S., Peng, J., Zhang, S., Li, J., Xiong, H., Zhang, W.:
  {Informer}: Beyond efficient {Transformer} for long sequence time-series
  forecasting. In: Proceedings of the AAAI Conference on Artificial
  Intelligence. vol.~35, pp. 11106--11115 (2021).
  \doi{10.1609/aaai.v35i12.17325}

\end{thebibliography}
\end{document}